\documentclass[11pt,a4paper]{article}
\usepackage[hyperref]{acl2021}
\usepackage{times}
\usepackage{latexsym}

\usepackage{microtype}

\usepackage{graphicx} 
\usepackage{amsmath}
\usepackage{subfigure}
\usepackage{booktabs}
\usepackage{multirow}
\usepackage{booktabs}
\usepackage{array}
\usepackage{bm}
\newcolumntype{L}[1]{>{\raggedright\let\newline\\\arraybackslash\hspace{0pt}}m{#1}}
\newcolumntype{C}[1]{>{\centering\let\newline\\\arraybackslash\hspace{0pt}}m{#1}}
\newcolumntype{R}[1]{>{\raggedleft\let\newline\\\arraybackslash\hspace{0pt}}m{#1}}

\aclfinalcopy 
\def\aclpaperid{568} 

\title{Diversifying Dialog Generation via Adaptive Label Smoothing}

\author{
Yida Wang\textsuperscript{1,2}\thanks{\quad Equal contribution},
Yinhe Zheng\textsuperscript{1,3}\footnotemark[1],
Yong Jiang\textsuperscript{2},
Minlie Huang\textsuperscript{1} \thanks{\quad Corresponding Author: aihuang@tsinghua.edu.cn} \\
\textsuperscript{1} The CoAI group, DCST, Institute for Artificial Intelligence, State Key Lab of Intelligent \\
Technology and Systems, Beijing National Research Center for Information \\
Science and Technology, Tsinghua University, Beijing, China \\
\textsuperscript{2} Tsinghua-Berkeley Shenzhen Institute, Tsinghua Shenzhen International Graduate School,
\\Tsinghua University, Shenzhen, China\\
\textsuperscript{3} Samsung Research China - Beijing (SRC-B)\\
\texttt{wangyd18@mails.tsinghua.edu.cn,} \texttt{yh.zheng@samsung.com,} \\
\texttt{jiangy@sz.tsinghua.edu.cn,} \texttt{aihuang@tsinghua.edu.cn} \\
}

\date{}

\begin{document}
\maketitle
\begin{abstract}
Neural dialogue generation models trained with the one-hot target distribution suffer from the over-confidence issue, which leads to poor generation diversity as widely reported in the literature. Although existing approaches such as label smoothing can alleviate this issue, they fail to adapt to diverse dialog contexts. In this paper, we propose an \textbf{Ada}ptive \textbf{Label} Smoothing (\textbf{AdaLabel}) approach that can adaptively estimate a target label distribution at each time step for different contexts. The maximum probability in the predicted distribution is used to modify the soft target distribution produced by a novel light-weight bi-directional decoder module. The resulting target distribution is aware of both previous and future contexts and is adjusted to avoid over-training the dialogue model. Our model can be trained in an end-to-end manner. Extensive experiments on two benchmark datasets show that our approach outperforms various competitive baselines in producing diverse responses. 
\end{abstract}

\section{Introduction}
\label{sec: intro}

The success of neural models has greatly advanced the research of dialog generation \cite{huang2020challenges,wang2020large,zhang2020dialogue}. However, most of these models suffer from a low-diversity issue where models tend to generate bland and generic responses such as \emph{I don't know} or \emph{I'm OK} \cite{li2015diversityMMI}. Although various approaches have been proposed to tackle this issue \cite{li2015diversityMMI,zhao2017learning,du2018variational,zhou2018elastic,welleck2019neural,zheng2020pre}, there are still remarkable gaps between responses generated by neural models and those from humans \cite{holtzman2019curious}. Further, some existing methods may even harm the fluency or coherence when improving the diversity of generated responses. \cite{ippolito2019comparison,massarelli2019decoding,zheng2021stylized}.

\begin{figure}[t]
\centering
\includegraphics[width=210px]{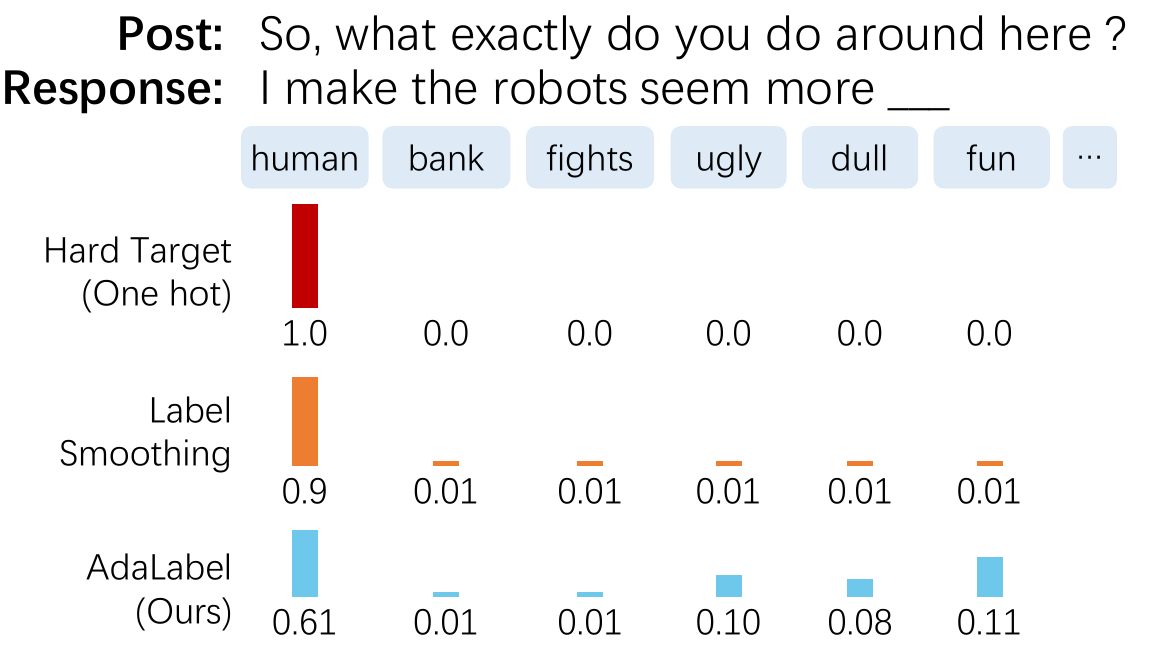} 
\caption{A dialogue sampled from the OpenSubtitles dataset. We demonstrate the hard target, label smoothing, and \textbf{Ada}ptive \textbf{Label} Smoothing approach when learning to predict the next word (\emph{``human''}). 
}
\label{fig:AST_samples}
\end{figure}

Recently, \citet{jiang2018sequence,jiang2019improving} show that there is a strong connection between the low-diversity problem and the over-confidence issue. i.e., over-confident dialogue models tend to produce low-diversity responses. One of the reasons can be attributed to the supervision target. Specifically, training a dialogue generation model with the Maximum Likelihood Estimation (MLE) objective under the hard target (i.e., one-hot distribution as ground truth) makes the model favor high-frequency tokens and produce over-confident probability estimation \cite{gowda2020neural}, which ultimately leads to poor calibration \cite{mukhoti2020calibrating}, and thus low diversity \cite{jiang2019improving}. \citet{hinton2015distilling} and \citet{yang2018knowledge} suggest that the ideal training target should be a soft target that assigns probability mass on multiple valid candidates (see Figure~\ref{fig:AST_samples}). With such a soft target, the over-confidence issue can be alleviated \cite{muller2019does}, and thus the diversity of the output responses can be improved.

Unfortunately, the ideal soft target is challenging to obtain. Early works try to tackle this issue using label smoothing \cite{szegedy2016rethinking}, i.e., a small probability is uniformly assigned to non-target words. However, the target distribution constructed in this way is far from ideal:
\textbf{First}, the probability of the target word is chosen manually and fixed, which cannot adapt to different contexts. However, as \citet{holtzman2019curious} demonstrated, human text distribution exhibits remarkable fluctuations in the per-token perplexity. We argue that different target probabilities should be used for different contexts.
\textbf{Second}, the uniform assignment of the probability mass on non-target words ignores the semantic relationship between the context and each word. Ideally, a word should receive more probability mass if it is more relevant to the context. For the example shown in Figure~\ref{fig:AST_samples}, word ``\emph{fun}'' is more likely to appear behind the context ``\emph{I make the robots seem more \_}'' than word ``\emph{bank}''.

To address the above issue, we propose an \textbf{Ada}ptive \textbf{Label} smoothing (AdaLabel) method that can dynamically estimate a soft target distribution at each time step for different contexts. Specifically, for each target word $y_t$ in the training data, the probability distribution predicted by the current model is first obtained. The maximum probability $p_{max}$ in this distribution measures the confidence of the current prediction, i.e., a higher $p_{max}$ means higher confidence for the current prediction. To avoid over-confidence, we use $p_{max}$ as the supervision signal for the target word $y_t$ in the training process so that the model will not be optimized towards $y_t$ when it correctly predicts $y_t$. A word-level factor is also introduced to facilitate the learning of low-frequency words. 

Moreover, we introduce a novel auxiliary decoder module $D_{a}$ to produce the supervision signals for these non-target words in each training step. $D_{a}$ only contains one transformer block, and it is optimized to predict words based on bi-directional contexts. A novel Target-Mask attention scheme is devised to prevent $D_{a}$ from seeing the target word in the training process. This scheme also enables parallel training and inference of $D_{a}$.

We perform extensive experiments on two benchmark datasets: DailyDialog and OpenSubtitles. Our method outperforms various competitive baselines and significantly improves the diversity of generated responses while ensuring fluency and coherency. Our major contributions are summarized:

\textbf{1.} We propose AdaLabel, a method that can produce a soft target distribution considering the current context and the model's confidence. Specifically, AdaLabel ensures that the dialogue model will not be optimized toward the target word $y_t$ if $y_t$ has been correctly predicted. This prevents our model from being over-confident.

\textbf{2.} We introduce a light-weight bi-directional decoder that can produce context-aware supervision signals for non-target words. A novel Target-Mask attention scheme is devised to facilitate the parallel training and inference of this decoder.

\textbf{3.} Extensive experiments on two benchmark dialogue datasets with both automatic and human evaluation results show that our method helps to alleviate the model over-confident issue and significantly improves the model's diversity.

\section{Related work}

\textbf{Diversity Promotion:} Existing approaches for solving the low diversity issue of neural dialogue models generally involve two categories:

The first category is training-based, where new training objectives are designed \cite{li2015diversityMMI,zhang2018generating,gao2019jointly} or latent variables are introduced \cite{zhao2017learning,zhou2018elastic} in the dialogue model. Some methods also try to refine the training target used in the MLE loss \cite{choi-etal-2020-f,jiang2019improving,li2019data}, or directly penalize the trivial responses with auxiliary loss terms \cite{welleck2019neural,li2019don}. Unlike these existing approaches, our method tries to adaptively adjust the training target by utilizing the current predictions.

The second category is decoding-based, in which different heuristic decoding rules are designed \cite{holtzman2019curious,kulikov2018importance}. Note that these decoding techniques are independent of the model setting, and our method can be used in combination with these techniques.

\begin{figure*}[!htp]
  \centering
  \includegraphics[width=310px]{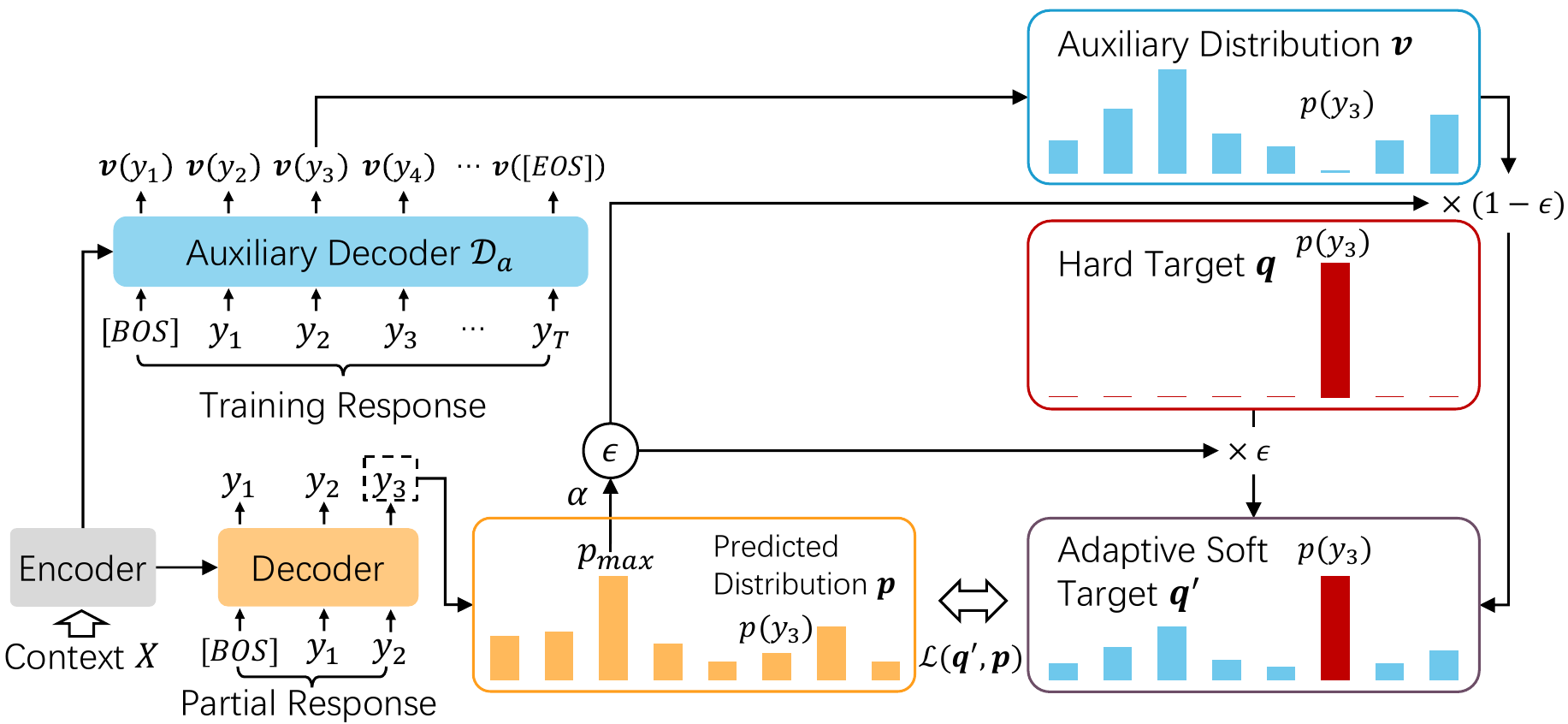}
  \caption{Overview of constructing the adaptive soft target $\bm{q'}$ using AdaLabel: The maximum probability $p_{max}$ in the predicted distribution $\bm{p}$ is used to obtain an adaption factor $\epsilon$, which is further used to combine the hard target $\bm{q}$ and the auxiliary distribution $\bm{v}$ to obtain $\bm{q'}$. A bi-directional auxiliary decoder $\mathcal{D}_a$ is used to produce $\bm{v}$.}
  \label{fig:overview}
\end{figure*}

\textbf{Confidence Calibration:} Modern deep neural networks suffer from the over-confidence issue \cite{guo2017calibration,kumar2019calibration}, and various remedies are proposed \cite{pereyra2017regularizing,mukhoti2020calibrating,lin2017focal}. Following the work of \citet{jiang2018sequence,jiang2019improving}, our method is proposed to tackle the over-confidence issue to improve the diversity of the generated responses. However, different from existing approaches, our method enables more flexible controls over the target distribution.

\textbf{Knowledge Distillation:} Another important technique similar to our work is knowledge distillation, in which a learned teacher model is distilled to a student model by minimizing a KL term \cite{hinton2015distilling,kim-rush-2016-sequence}.

The most related work comparing to ours is the C-MLM approach \cite{chen2020distilling}, in which a BERT model is fine-tuned to be a teacher. Our approach and C-MLM's primary difference is that our auxiliary decoder $D_a$ is a one layer module that is jointly trained with the dialogue model. However, the BERT teacher in C-MLM contains much more parameters, and it is trained using an expensive pre-trained and then fine-tuned process. Moreover, the target-masked attention scheme in $D_a$ enables parallel inferences of $\bm{v}$ for each training sequence $Y$. In contrast, multiple independent forward passes are required for the BERT teacher.

\section{Method}

\subsection{Background: MLE with Hard Target}

The goal of generative dialogue modeling is to learn a conditional probability distribution $p(Y|X)$, where $X$ is the dialogue context, $Y=y_1, ..., y_T$ is a response word sequence, and $y_i \in \mathcal{V}$ is a word from the vocabulary $\mathcal{V}$. In an auto-regressive manner, $p(Y|X)$ is factorized as $\prod_{t}p(y_t|y_{<t},X)$. For each target word $y_t$ in the training sequence $Y$, a conventional MLE training approach try to optimize the following cross entropy loss:
\begin{equation}\label{eq:MLE_loss}
    \mathcal{L}(\bm{q}, \bm{p})=-\sum_{w_k \in \mathcal{V}} q_k {\rm log} \left[ p(w_{k}|y_{<t}, X) \right],
\end{equation} 
where $\bm{q}$ is a one-hot distribution (i.e., a hard target) that assigns a probability of $1$ for the target word $y_t$ and 0 otherwise, i.e., $q_k=1$ only when $w_k = y_t$. For simplicity of notation, we abbreviate the dependency of $y_t$ in the notation of each distribution in our paper, i.e., different target word $y_t$ in $Y$ corresponds to different values of $\bm{q}$ and $\bm{p}$.

\subsection{Method Overview}

We propose to adaptively construct a soft target distribution $\bm{q'}$ to replace $\bm{q}$ in Eq. \ref{eq:MLE_loss}. Specifically,
\begin{equation} \label{eq:soft_tgt_dis}
    \bm{q'} = \varepsilon \cdot \bm{q} + (1-\varepsilon) \cdot \bm{v},
\end{equation}
where $\varepsilon \in [0, 1]$ is an adaption factor, and $\bm{v}$ is an auxiliary distribution vector that depends on the current time step. (see Figure \ref{fig:overview} for an overview). 

In this study, we constrain $\bm{v}$ to assign zero probability for the target word $y_t$ and non-zero probabilities for these non-target words $\mathcal{V}_{\neq y_t}=\{y_i | y_i \in \mathcal{V}, y_i \neq y_t\}$. This constraint allows us to explicitly control the supervisions assigned to $y_t$. Specifically, the first term $\varepsilon \cdot \bm{q}$ and the second term $(1-\varepsilon) \cdot \bm{v}$ in Eq. \ref{eq:soft_tgt_dis} respectively determines how much probability $\bm{q'}$ assigns to $y_t$ and $\mathcal{V}_{\neq y_t}$. This setting differs from conventional knowledge distillation \cite{kim-rush-2016-sequence} because it facilitates more flexible controls over $\bm{q'}$, so that we can use the factor $\varepsilon$ to determine the supervision signal provided for the target word $y_t$. The following sections detail how to compute $\varepsilon$ and $\bm{v}$.

\subsection{Target Word Probability}\label{sec:target_word_prob}

We control the probability of the target word $y_t$ in $\bm{p'}$ by manipulating the adaption factor $\varepsilon$ in Eq. \ref{eq:soft_tgt_dis}. Specifically, for a training dialogue pair $\langle X,Y \rangle$ and each target word $y_t \in Y$, the current distribution $\bm{p}(\cdot | y_{<t}, X)$ is first calculated, and the maximum probability in this distribution is obtained:
\begin{equation} \label{eq:p_max}
    p_{max} = \max_{w_k \in \mathcal{V}} p(w_k | y_{<t}, X).
\end{equation}
$\varepsilon$ is then obtained:
\begin{align} \label{eq:epsilon}
    \varepsilon &= \textrm{max}(p_{max}, \lambda),
\end{align}
where $\lambda$ serves as a lower-bound of $\varepsilon$ (i.e., $\varepsilon \geq \lambda$). 

The basic intuition behind Eq. \ref{eq:epsilon} is to set $\varepsilon = p_{max}$ when $p_{max}$ is reasonably large. This design prevents our model from receiving supervisions sharper than $p_{max}$, when the current prediction is confidence enough.  

Further, to ensure that the target word $y_t$ always receives the largest probability in $\bm{q'}$, i.e., to ensure $\varepsilon > (1-\varepsilon) \cdot \textrm{max}(\bm{v})$ (see Eq. \ref{eq:soft_tgt_dis}), in which $\textrm{max}(\bm{v})$ is the maximum probabilities for non-target words $\mathcal{V}_{\neq y_t}$, we have to enforce $\varepsilon > \frac{\textrm{max}(\bm{v})}{1+\textrm{max}(\bm{v})}$. Thus we propose to calculate the lower-bound $\lambda$ of $\varepsilon$ as:
\begin{align}\label{eq:lamda}
\lambda =\frac{\textrm{max}(\bm{v})}{1+\textrm{max}(\bm{v})} + \eta, 
\end{align}
where $\eta > 0$ is a hyper-parameter that controls the margin between the probability of the target word and non-target words in $\bm{p'}$.

To facilitate faster converge and better learning of low-probability words, an empirical factor $\alpha \in [0, 1]$ is further introduced to adjust the calculation of $\varepsilon$ on the basis of Eq. \ref{eq:epsilon}:
\begin{align}\label{eq:epsilon_refine}
    \varepsilon = 1 - \alpha \cdot (1 - \textrm{max}(p_{max}, \lambda)),
\end{align}
where $\alpha$ is calculated as the relative ratio to $p_{max}$:
\begin{align}\label{eq:alpha}
    \alpha = \left[ \frac{p(y_t| y_{<t}, X)}{p_{max}} \right]^2,
\end{align}
where $p(y_t| y_{<t}, X)$ is the probability for the target word $y_t$. Note that Eq. \ref{eq:epsilon_refine} and Eq. \ref{eq:epsilon} is equivalent if $\alpha=1$. Intuitively, $\alpha$ accelerates the training of low-frequency words because if $y_t$ is of low-frequency in the corpus, then $y_t$ is usually under-trained and thus $p(y_t| y_{<t}, X)$ is generally small. This leads to a small $\alpha$ and thus increases the probability for $y_t$ in $\bm{p'}$.

Note that $\varepsilon$, $\lambda$ and $\alpha$ are all time-step specific variables, whereas $\eta$ is a fixed hyper-parameter. This allows the values adapt to dynamic contexts. 
In our experiments, Eq. \ref{eq:epsilon_refine} is used to calculate $\varepsilon$.

\begin{figure}[!t]
  \centering
  \includegraphics[width=200px]{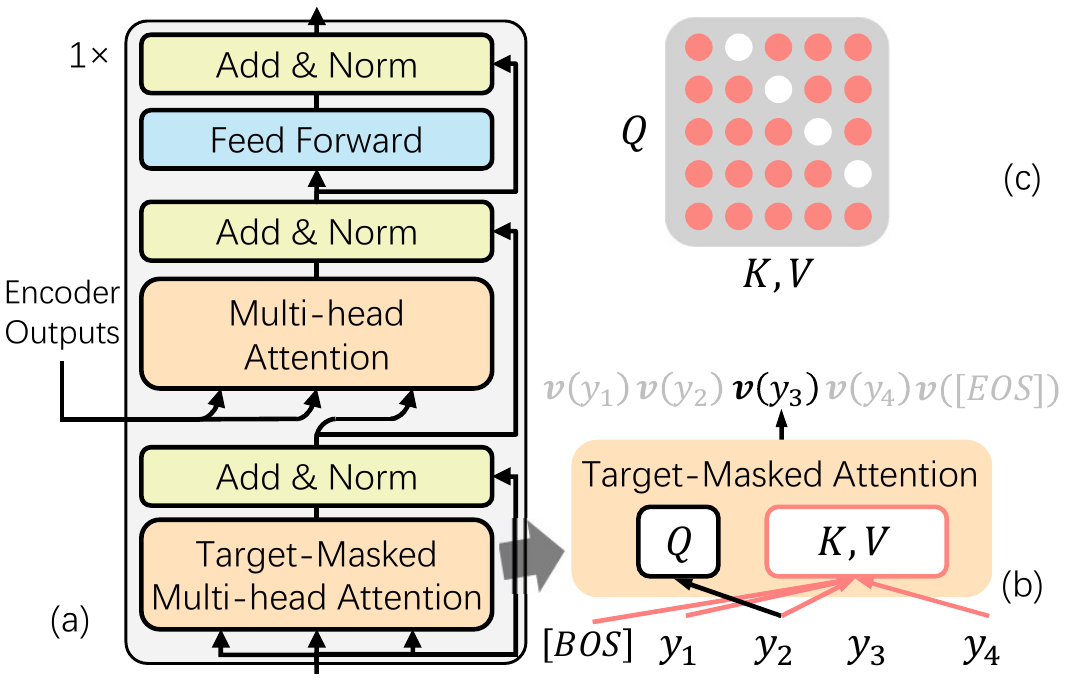}
  \caption{(a) The auxiliary decoder $D_a$; (b) The target-masked attention scheme used to compute the auxiliary distribution $\bm{v}$ for the target word $y_3$, specifically, $y_2$ is used as the query and $y_3$ is masked; (c) The attention pattern used in the target-masked attention scheme, white dots represent masked positions.}
  \label{fig:auxiliary_dec}
\end{figure}

\subsection{Non-target Words Probabilities}

The auxiliary distribution $\bm{v}$ in Eq. \ref{eq:soft_tgt_dis} is calculated using an auxiliary decoder $D_a$, which is a single-layer transformer-based decoder that is jointly optimized with the generation model. Figure \ref{fig:auxiliary_dec} shows the structure of $D_a$, in which a novel target-masked attention scheme is devised to mask each target word $y_t$ in the self attention module of the decoder when calculating the corresponding $\bm{v}$ (see Figure \ref{fig:auxiliary_dec}b and \ref{fig:auxiliary_dec}c). In this way, bi-directional contexts can be utilized when predicting the auxiliary distribution $\bm{v}$ for $y_t$. Moreover, it is important to use only one decoder layer in $D_a$ because stacking multiple layers in $D_a$ leaks the information of $y_t$ to $\bm{v}$.

Note that using one layer in $D_a$ does not necessarily downgrade its performance \cite{kasai2021deep}. Our experiment results in Section \ref{sec:aux_decoder} indicate that with the help of bi-directional contexts, the accuracy of $D_a$ largely outperforms the uni-directional dialogue decoder that is much deeper than $D_a$. Moreover, for a training response $Y$, the structure of $D_a$ enables us infer the auxiliary distribution in parallel for all the target words in $Y$ within a single forward pass. This differs from the BERT teacher used by \citet{chen2020distilling}, in which multiple independent forward passes are needed to get the teacher distributions for all the words in $Y$.

When training $D_a$, the following standard MLE loss is optimized for each target word $y_t$:
\begin{equation}\label{eq:aux_dec_loss}
    \mathcal{L}(\bm{q}, \bm{v}) = -\sum_{k=1}^{|\mathcal{V}|} q_k {\rm log} {v}_{k},
\end{equation}
in which the notation of $q_k$ follows Eq. \ref{eq:MLE_loss}. 

The outputs of $D_a$ are used as the logits to infer $\bm{v}$ to be further used in Eq. \ref{eq:soft_tgt_dis}. Specifically, the logit of the target word $y_t$ is masked to $-\infty$ before Softmax to ensure $y_t$ always receives zero probability in $\bm{v}$. Moreover, we also follow the approach used by \citet{tang2020understanding} to truncate the head and tail of the remaining logits before inferring $\bm{v}$ in Eq. \ref{eq:soft_tgt_dis}, i.e., all the logits are ranked in a descending order and only the logits ranked from $n$ to $m$ are kept while the rest logits are masked to $-\infty$. This masks the head and tail probabilities in $\bm{v}$ to zero. We argue that truncating the tail probabilities of $\bm{v}$ filters noises, and truncating the head probabilities of $\bm{v}$ encourages the dialogue model to focus more on low-probability words. In our experiments, we set $n=2$ and $m=500$. An extensive hyper-parameter search indicates that our method is not sensitive to the value of $n$ and $m$.

There are two major differences between our auxiliary decoder $D_a$ and the teacher model used in conventional knowledge distillation approaches: First, conventional teacher models usually carry more parameters than their students, whereas $D_a$ is rather light-weight. Second, conventional teacher models are typically pre-trained before being utilized in the distillation process, whereas $D_a$ is trained jointly with our dialogue model.

\section{Experiments}

\subsection{Dataset}
We use two benchmark datasets for open-domain dialogue generation:
\textbf{DailyDialog} \cite{li2017dailydialog} is a high-quality multi-turn dialogue dataset that is collected from daily conversations.
\textbf{OpenSubtitles} \footnote{\url{http://opus.nlpl.eu/OpenSubtitles.php}} contains dialogues collected from movie subtitles.
Moreover, we follow \citet{li2015diversityMMI} and \citet{jiang2019improving} to focus on short conversations, i.e., dialogues with posts or responses longer than 100 tokens are removed. See Table \ref{table:data_stat} for more details.

\begin{table}[!t]
\centering
\begin{tabular}{lccc}
\toprule
 & Train & Valid & Test\\
\midrule
DailyDialog   & 65.8K & 6.13K & 5.80K \\
OpenSubtitles & 1.14M & 20.0K & 10.0K \\
\bottomrule
\end{tabular}
\caption{Dataset statistics.}
\label{table:data_stat}
\end{table}

\subsection{Implementation Details}

The backbone of our model is the transformer-based sequence to sequence model \cite{vaswani2017attention}, and most hyper-parameters follow \citet{cai2020data}. Specifically, the encoder and decoder each contains 6 layers. Each layer has 8 attention heads, and the hidden size is set to 512. The auxiliary decoder $D_a$ follows the same hyper-parameter setting as the dialogue decoder, but it only contains one layer. The WordPiece tokenizer provided by BERT \cite{devlin2018bert} is used, and the Adam optimizer \cite{kingma2014adam} is employed to train our model from random initializations with a learning rate of 1e-4. $\eta$ in Eq. \ref{eq:lamda} is set to 0.2 for all datasets. See Appendix A for more details. \footnote{Our code is available at: \url{https://github.com/lemon234071/AdaLabel}}

\begin{table*}[!ht]
\centering
\small
\setlength\tabcolsep{2.2pt} 
\begin{tabular}{L{39pt}C{21pt}C{21pt}C{21pt}C{21pt}C{21pt}C{21pt}C{21pt}C{21pt}C{21pt}C{21pt}C{21pt}C{21pt}C{21pt}
C{21pt}C{21pt}C{21pt}}
\toprule
\multirow{2}{*}{Model} 
& \multicolumn{8}{c}{ DailyDialog } 
& \multicolumn{8}{c}{ OpenSubtitles } \\
\cmidrule(lr){2-9} \cmidrule(lr){10-17}
    & \multicolumn{2}{c}{Dist-1, 2} & \multicolumn{2}{c}{Ent-1, 2} & \multicolumn{1}{c}{LF} & \multicolumn{3}{c}{BLEU-2,3,4} 
    & \multicolumn{2}{c}{Dist-1, 2} & \multicolumn{2}{c}{Ent-1, 2} & \multicolumn{1}{c}{LF} & \multicolumn{3}{c}{BLEU-2,3,4} \\
    \midrule

      CE 
    &    1.67  &    9.43  &    4.53  &    6.59  &    2.99  &    7.56  &    4.38  &    2.61
    &   2.55 &   9.87 &   4.13 &   5.58 &   0.84 &   7.60 &   4.30 &   2.57\\ 
    
      LS 
    &    1.48  &    8.78  &    4.48  &    6.55  &    2.44  &    7.98  &    4.68  &    2.86
    &   2.77 &   13.08 &   4.45 &   6.57 &   0.51 &   8.91 &   5.57 &   3.84\\
    
      FL 
    &    2.38  &    13.42  &    4.7  &    7.04  &    5.05  &    9.74  &    6.12  &    4.11
    &   3.19 &   13.16 &   4.42 &   6.50 &   1.04 &   8.06 &   4.79 &  3.08\\
    
      FACE 
    &   1.62  &  11.04  &  4.96 &  7.27  &   4.11 &   8.78 &  5.06 &  3.06
    &   3.31 &   14.06 &   4.77 &   7.05 &   1.33 &   7.69 &  4.40 &  2.70 \\
    
      F$^2$ 
    &   1.40 &   7.91 &   4.35 &   6.28  &  2.32  &  7.78 &   4.45 &  2.60
    &   2.89 &  11.40 &   4.24 &   6.14 &   0.99 &   7.52 &   4.30 &  2.62\\
    
      CP 
    &    2.35  &    12.91  &    4.64  &    6.89  &    4.07  &    9.06  &    5.68  &    3.79
    &   3.11 &   12.72 &   4.36 &   6.35 &    0.98 &   8.06 &    4.82 &   3.12 \\
    
      UL 
    &    2.35  &    12.99  &    4.68  &    6.98  &    4.96  &    10.83  &    6.87  &    4.61
    &   2.84 &   11.64 &   4.31 &   6.32 &    0.76 &    7.73 &   4.59 &   2.96\\
    
      NL 
    &    1.66  &    9.18  &    4.47  &    6.58  &    4.30  &    9.83  &    5.83  &    3.60
    &   3.24 &   12.98 &   4.42 &   6.49 &    1.08 &    7.56 &   4.38 &  2.71\\
    
      D2GPo 
    &    1.26  &    8.06  &    4.43  &    6.48  &    2.20  &    8.30  &    4.82  &    2.93
    &   2.07 &  11.01 &   4.32 &   6.36 &   0.19 &    8.41 &   5.08 &  3.35 \\
    
    AdaLabel 
    &   \textbf{3.96} &   \textbf{23.53} &   \textbf{5.17} &   \textbf{8.00} &   \textbf{8.49} &    \textbf{17.42} &   \textbf{13.38} &  \textbf{11.01} 
    &   \textbf{4.78} &   \textbf{22.88} &   \textbf{4.96} &   \textbf{7.66} &   \textbf{1.47} &    \textbf{9.80} &   \textbf{6.48} &  \textbf{4.75} \\
    
\midrule
    Human 
    &   6.59 &   37.74 &   5.67 &   8.91  &   13.7 &   N/A &   N/A &   N/A
    &   8.62 &   43.16 &   5.89 &   9.36  &   4.75 &   N/A &   N/A &   N/A  \\
    \bottomrule
\end{tabular}
\caption{Automatic evaluation results (\%). Best results among all the models are in bold.}
\label{tab:auto_res}
\end{table*}

\subsection{Baselines}

We compared our method with two groups of baselines that try to tackle the over-confidence issue.

The first group modifies the training target used to compute the loss function:
\textbf{1) LS} \cite{szegedy2016rethinking}: uses the \underline{l}abel \underline{s}moothing approach to construct a target distribution by adding the one-hot target and a uniform distribution;
\textbf{2) FL} \cite{lin2017focal}: uses the \underline{f}ocal \underline{l}oss to down-weigh well-classified tokens in each time step.
\textbf{3) FACE} \cite{jiang2019improving}: uses the \underline{f}requency-\underline{a}ware \underline{c}ross-\underline{e}ntropy loss to balance per-token training losses. Specifically, relative low losses are assigned to high-frequency words to explicitly tackle the over-confidence issue. We used the best performing ``Pre-weigh'' version in our experiments.
\textbf{4) F$^{\bm{2}}$} \cite{choi-etal-2020-f}: factorizes the target distribution based on the token frequencies.

The second group of baselines add some penalty term to the standard MLE loss:
\textbf{5) CP} \cite{pereyra2017regularizing}: a \underline{c}onfidence \underline{p}enalty term is added to regularize the entropy of the model, so that over-confident predictions are penalized;
\textbf{6) UL} \cite{welleck2019neural}: an \underline{u}nlikelihood \underline{l}oss term is added to penalize the frequently generated words.
\textbf{7) NL} \cite{he2019negative}: works similarly with baseline \textbf{UL} except a \underline{n}egative \underline{l}oss term is used instead of the unlikelihood loss term.
\textbf{8) D2GPo} \cite{li2019data}: augments the MLE loss with a \underline{d}ata-dependent \underline{g}aussian \underline{p}rior \underline{o}bjective to assign different losses for different non-target words.

We also compared to: \textbf{9) CE}: a vanilla Seq2Seq model trained with the \underline{c}ross-\underline{e}ntropy loss. For fair comparisons, the C-MLM model proposed by \citet{chen2020distilling} is not used as our baseline since the BERT teacher in C-MLM requires a large amount of extra data to pre-train. Nevertheless, AdaLabel still surpasses C-MLM on various metrics (see Appendix F for more analysis).

All our baselines are adapted from the authors' official codes with the same backbone architecture and hyper-parameters as our model (see details in Appendix B). Following the original setting, a train-and-refine strategy is used in baseline 3, 6, and 7, i.e., these baselines are refined based on \textbf{CE}. We follow the setting of \citet{jiang2019improving} to use deterministic decoding scheme (particularly, greedy decoding) for our model and all baselines. Note that our method can be adapted to other decoding schemes such as beam-search or top-K sampling. See Appendix C for more detailed analysis.

\subsection{Automatic Evaluation}
\textbf{Metrics:}
We first used automatic metrics to evaluate our method:
1) \emph{Distinct} (\textbf{Dist}) \cite{li2015diversityMMI} calculates the proportion of unique n-grams (n=1, 2) in the generated responses, which is widely used to measure the response diversity.
2) \emph{Entropy} (\textbf{Ent}) \cite{zhang2018generating} evaluates how evenly the empirical n-gram (n=1, 2) distribution is. Higher sores mean more diverse of the response.
3) \emph{Low-Frequency Token Ratio} (\textbf{LF}) \cite{li2019data} further measures the model diversity by counting the ratio of low-frequency words in the generated responses. We chose words with a frequency less than 100 in each corpus as low-frequency words. Over-confident models tend to omit low-frequency words (i.e., get low \textbf{LF} scores) and yield less diversified responses.
4) \emph{\textbf{BLEU}} \cite{papineni2002bleu} measures n-gram (n=2, 3, 4) overlap between the generated responses and references.

\textbf{Results:}
As shown in Table \ref{tab:auto_res}, our method AdaLabel outperforms all the baselines by large margins on all the datasets. We can further observe that: 1) AdaLabel achieves the best diversity scores (Dist-1,2, Ent-1,2, and LF). This indicates that our method yields better training targets that help to produce more diverse responses; 2). The models that explicitly tackle the over-confidence issue (i.e., AdaLabel and FACE) generally outperform other baselines in diversity-related metrics. For example, FACE obtains the second-best diversity scores (i.e., Dist, Ent, and LF) on the OpenSubtitles dataset. This verifies our motivation that alleviating the over-confidence issue helps to produce more diverse responses.

Note that our method also outperforms all the baselines using the stochastic decoding scheme. Please refer to Appendix C for more details.

\subsection{Manual Evaluation} \label{sec: human}

\textbf{Metrics:}
Pairwise manual evaluations are conducted to further validate our method. Specifically, for a given dialogue post, our model's response is paired with the one from a baseline. Three individual annotators were employed to rank each response pair from three aspects:
1) \emph{Fluency} (\textbf{Flu.}): which response is more fluent;
2) \emph{Coherency} (\textbf{Coh.}): which response is more coherent to the context;
3) \emph{Informativeness} (\textbf{Info.}): which response contains more informative content.
We also asked the annotator to choose an overall preferred response (\textbf{Pref.}). Ties were allowed.

\textbf{Results:}
200 posts were randomly sampled from each of these two datasets, respectively, and totally 3.6K response pairs were generated. The inter-rater annotation agreement was measured using Fleiss's kappa $\kappa$ \cite{fleiss1971measuring}. Particularly, the $\kappa$ value on DailyDialog, OpenSubtitles dataset was 0.59 and 0.55, respectively, indicating moderate agreement.

\begin{table*}[t]
\small
\setlength\tabcolsep{3.2pt} 
\centering
    \begin{tabular}{L{80pt}C{32pt}C{32pt}C{32pt}C{32pt}C{32pt}C{32pt}C{32pt}C{32pt}}
    \toprule
    \multirow{2}{*}{Comparison} & \multicolumn{4}{c}{DailyDialog} & \multicolumn{4}{c}{OpenSubtitles} \\
    \cmidrule(lr){2-5} \cmidrule(lr){6-9} 
    
    & 
    Pref. & Flu. & Coh. & Info. & 
    Pref. & Flu. & Coh. & Info. \\
    
    \midrule
    
      AdaLabel vs CE 
    & 17.00$^\ddagger$ & 1.33  & 12.5$^\ddagger$ & 28.33$^\ddagger$ & 
      6.33             & 1.17  & 7.33$^\dagger$  & 13.67$^\ddagger$ \\

      AdaLabel vs LS 
    & 2.67             & 0.17  & 3.33 & 24.83$^\ddagger$ &  
      5.3              & -0.67 & 3.17 & 8.50$^\ddagger$  \\
    
      AdaLabel vs FL 
    & 4.50             & 1.67  & 7.00$^\dagger$   &   22.0$^\ddagger$  &  
      8.00$^\dagger$   & 1.00  & 6.00             &   5.50             \\
      
      AdaLabel vs FACE 
    & 6.67$^\dagger$ &   3.50$^\dagger$ &   7.17$^\dagger$ &   8.50$^\dagger$   &  
      4.50         &   0.50           &   1.83          &   2.50             \\
      
      AdaLabel vs F$^2$
    & 7.67$^\dagger$ &   0.33             &   6.83$^\dagger$ &   8.67$^\ddagger$  &
      4.33           &   -0.50            &   1.67           &   9.50$^\ddagger$    \\
    
      AdaLabel vs CP 
    & 10.50$^\ddagger$ & -0.17          &   8.00$^\dagger$   & 23.83$^\ddagger$ &  
      8.00$^\dagger$   & 1.50           &   6.17             & 16.83$^\ddagger$ \\

      AdaLabel vs UL 
    & 7.83$^\dagger$ &   0.83 &   6.67$^\dagger$ &   17.33$^\ddagger$ &  
      6.83$^\dagger$ &   2.00 &   5.83           &   15.00$^\ddagger$ \\
    
      AdaLabel vs NL 
    & 9.17$^\dagger$   &   2.67$^\dagger$ &   9.17$^\dagger$  &   7.67$^\dagger$  &  
      5.17             &   0.17           &   2.17            &   15.5$^\ddagger$ \\
      
      AdaLabel vs D2GPo 
    & 0.83             &   0.00          &   3.33            &   15.17$^\ddagger$&
      3.17             &   7.33$^\ddagger$&   1.00            &   6.33$^\dagger$   \\
    
    \bottomrule
    \end{tabular}
\caption{Pairwise human evaluation results (\%). The absolute gains of AdaLabel (i.e., \emph{Win rate} $-$ \emph{Lose rate}) are reported. $^\dagger$, $^\ddagger$ indicates significant improvement with \emph{p}-value $<0.05$ and $<0.005$, respectively (sign test).}
\label{tab:manual_res}
\end{table*}

As shown in Table \ref{tab:manual_res}, AdaLabel outperforms all the baselines on the informativeness measure. This means that our method can respond with more informative content. We can further observe that:

1). All models achieve competitive fluency because it is easy for neural models to produce fluent responses by yielding trivial responses like ``I don't know''. However, our model surpasses most baselines in terms of fluency while ensuring high diversity scores. This demonstrates the superiority of our method in producing high quality responses.

2). AdaLabel produces more coherent responses comparing to most baselines. This verifies that our model does not sacrifice the response quality when achieving high diversity scores. In fact, by controlling the model's confidence, more low-frequency words are encouraged, and thus AdaLabel can produce more relevant and coherent responses. This claim is further verified by observing that our model achieves the best overall preference score among all the baselines.

\subsection{Ablation study}

\begin{table}[t]
\small
\setlength\tabcolsep{2.0pt}
\centering
  \begin{tabular}{lC{21pt}C{21pt}C{21pt}C{21pt}C{21pt}C{21pt}C{21pt}}
  \toprule
    Model &  \multicolumn{2}{c}{BLEU-3,4}& \multicolumn{2}{c}{Dist-1,2} & \multicolumn{2}{c}{Ent-1,2} & LF   \\
  \midrule
    1.w/o $\varepsilon$ &   5.46 &   3.57 &   2.52 &   13.21 &   4.64 &   6.89 &   4.85 \\
    2.w/o $\alpha$      &  11.35 &   8.70 &   3.62 &   20.56 &   5.02 &   7.70 &   7.30 \\
  \midrule
    3.Orig. $\bm{v}$    &   8.15 &   5.77 &   3.71 &   19.53 &   5.00 &   7.58 &   8.25 \\
    4.Uniform           &   5.66 &   3.61 &   2.24 &   14.96 &   4.84 &   7.33 &   4.98 \\
    5.Rand              &   6.27 &   4.07 &   2.03 &   13.47 &   4.7 &   7.08 &   4.56 \\
    6.BERT              &   11.6 &   9.34 &   3.67 &   20.97 &   5.02 &   7.71 &   7.28 \\
  \midrule
    AdaLabel &  \textbf{13.38} &   \textbf{11.01}     &    \textbf{3.96} &   \textbf{23.53} &   \textbf{5.17} &   \textbf{8.00} &   \textbf{8.49} \\
  \bottomrule
  \end{tabular}
\caption{
Ablation study results on DailyDialog (\%). 
}
\label{tab:ablation}
\end{table}

Ablation studies were performed to verify the effect of each component in our method. Specifically, two groups of variants were tested:

The first group validates the effectiveness of the calculated target word probability, i.e., $\varepsilon$:
\textbf{1). w/o $\bm{\varepsilon}$} directly sets a fixed value for $\varepsilon$ in Eq. \ref{eq:soft_tgt_dis}. The specific value of $\varepsilon$ is searched from 0.1 to 0.7 with a stride of 0.1;
\textbf{2). w/o $\bm{\alpha}$} omits the empirical factor $\alpha$ in calculating $\varepsilon$, i.e., the value of $\varepsilon$ in Eq. \ref{eq:soft_tgt_dis} is calculated using Eq. \ref{eq:epsilon} in instead of Eq. \ref{eq:epsilon_refine}.

The second group validates the effectiveness of the non-target word probabilities produced by $D_a$, i.e., $\bm{v}$: 
\textbf{3). Orig. $\bm{v}$} does not truncate the head of $\bm{v}$ when inferring from $D_a$. Note that the truncation for the tail of $\bm{v}$ is still applied since its effectiveness has already been proved in previous studies \cite{tang2020understanding,tan2019multilingual};
\textbf{4). Uniform} uses an uniform distribution as $\bm{v}$ in Eq. \ref{eq:soft_tgt_dis}. Note that different from the baseline \textbf{LS}, the value of $\varepsilon$ is calculated using Eq. \ref{eq:epsilon_refine} in this ablation model, whereas the value of $\varepsilon$ in the baseline \textbf{LS} is fixed ;
\textbf{5). Rand} use a random distributions as $\bm{v}$ in Eq. \ref{eq:soft_tgt_dis};
\textbf{6). BERT} follows the work of \citet{chen2020distilling} to fine-tune a pre-trained BERT model to produce $\bm{v}$.
Note that our dialogue model may benefit from the multi-task training of $D_a$ since $D_a$ shares the same encoder with our dialogue model. Optimizing Eq. \ref{eq:aux_dec_loss} may help the encoder to capture better features. For fair comparison, we kept the task of optimizing $D_a$ in ablation models 4-6 although it is not used to infer $\bm{v}$.

Table \ref{tab:ablation} shows the results of ablation models on the DailyDialog dataset. As can be seen from the first two rows, our method to adaptively calculate $\varepsilon$ helps to improve the performance of our model by a large margin, and the empirical adjustment factor $\alpha$ helps to further improve our performance by facilitating the learning of low-probability words. The performance of ablation models 3-6 in Table \ref{tab:ablation} proves that $\bm{v}$ captures reliable distribution and helps our model produce more diverse responses. Moreover, truncating the head distribution of $\bm{v}$ enables the dialogue model to focus more on the low-frequency words and thus facilitates more informative responses. 

It is also interesting to note that our auxiliary decoder $D_a$ surpasses the BERT teacher used by \citet{chen2020distilling} in helping the dialogue model to produce more diverse responses. This further proves the effectiveness of $D_a$ considering that BERT contains 6 times parameters than $D_a$ and consumes much more computation resources.

\begin{table}[t]
\small
\setlength\tabcolsep{2.0pt}
\centering
\begin{tabular}{lcc}
\toprule
 & DailyDialog & OpenSubtitles \\
\midrule
Auxiliary Decoder $D_a$  & \textbf{64.03} & \textbf{64.92} \\
Dialog Decoder in AdaLabel & 44.16 & 43.90 \\
Dialog Decoder in CE         & 38.58 & 41.57 \\
\bottomrule
\end{tabular}
\caption{Prediction accuracy of decoders on test sets.}
\label{tab:acc}
\end{table}

\begin{figure}[!t]
  \centering
  \includegraphics[width=200px]{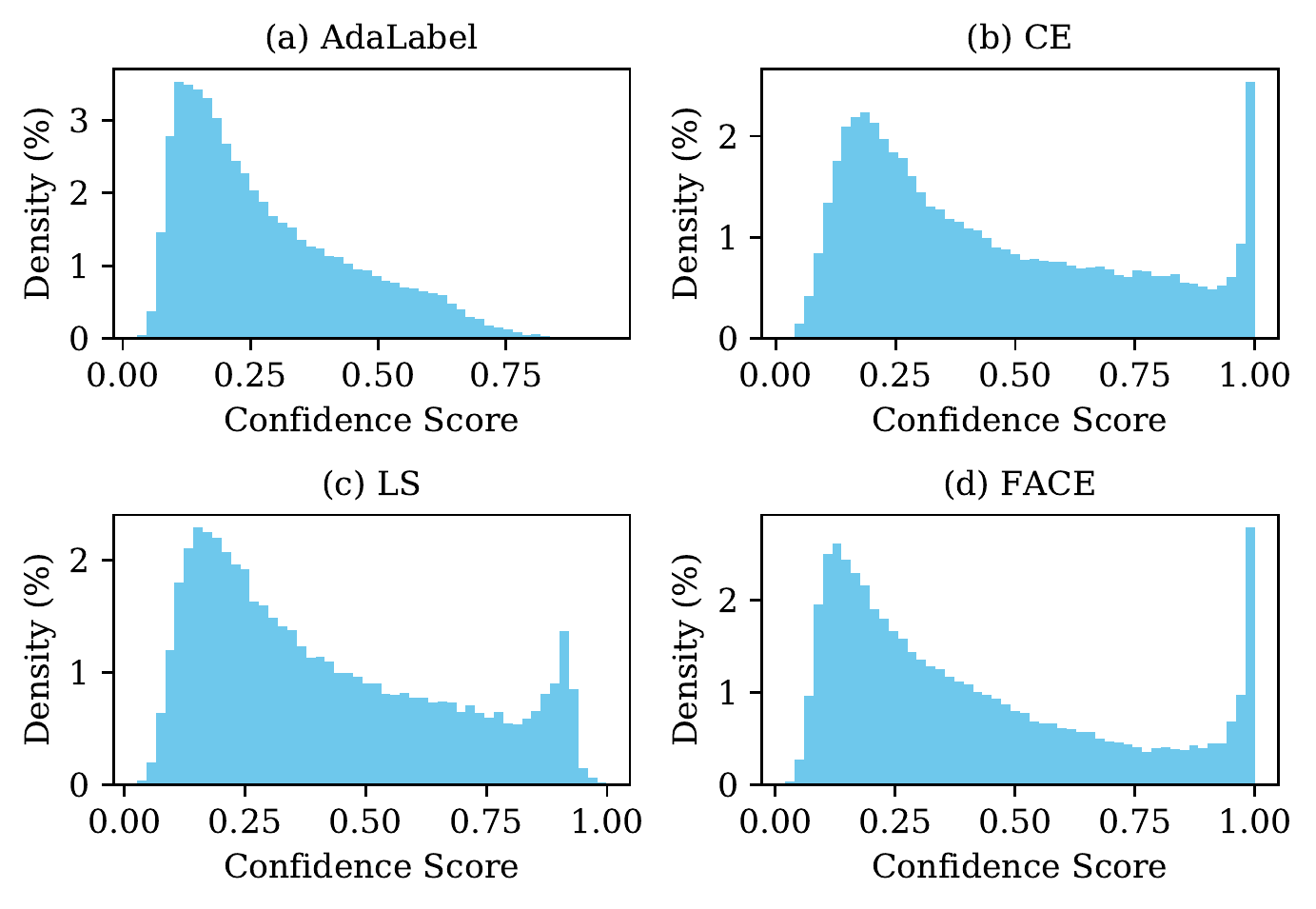}
  \caption{Empirical distribution of confidence scores for high-frequency words on the OpenSubtitles dataset. Words occupying the top 40\% of the frequency mass in the training set are regarded as high-frequency words.}
  \label{fig:conf_dist}
\end{figure}

\section{Discussion}

\subsection{Auxiliary Decoder}\label{sec:aux_decoder}
To further test the performance of $D_a$, we evaluated the averaged accuracy score of $D_a$ when predicting each target word in the test set (first row in Table \ref{tab:acc}). Specifically, a target word $y_t$ in the reference response is determined to be correctly predicted if it is top-ranked in the predicted distribution $p(\cdot | y_{<t},X)$. A better decoder is generally believed to obtain a higher accuracy. Table \ref{tab:acc} also reports the uni-directional dialogue decoders' accuracy in AdaLabel and CE. It can be seen that $D_a$ can make substantially more accurate predictions with the help of modeling bi-directional contexts using only one layer. Moreover, the dialogue model's decoder in AdaLabel, which is guided by $D_a$, achieves better accuracies than the CE. This further proves that our light-weight $D_a$ is capable of producing effective $\bm{v}$.

\subsection{Prediction Confidence}

We also visualized the distribution of confidence scores assigned by each dialogue model to high-frequency words. 
Figure \ref{fig:conf_dist} shows the results of four best performing models on the OpenSubtitles dataset. The spikes of high confidence score observed in Figure \ref{fig:conf_dist}b and \ref{fig:conf_dist}d indicate that CE and FACE assign extremely high confidence scores to a large number of high-frequency words. Although the smoothed labels in LS manage to alleviate these high-confidence-spikes (Figure \ref{fig:conf_dist}c), a considerable amount of words still receives high confidence scores in LS. Our model outperforms all the baselines to avoid assigning over-confidence scores, thus alleviating the over-confidence issue. A similar trend is also observed on the DailyDialog dataset (see Appendix D for results of all models on both datasets).

\subsection{Predicted Rare Word Distribution}

\begin{figure}[!t]
  \centering
  \includegraphics[width=210px]{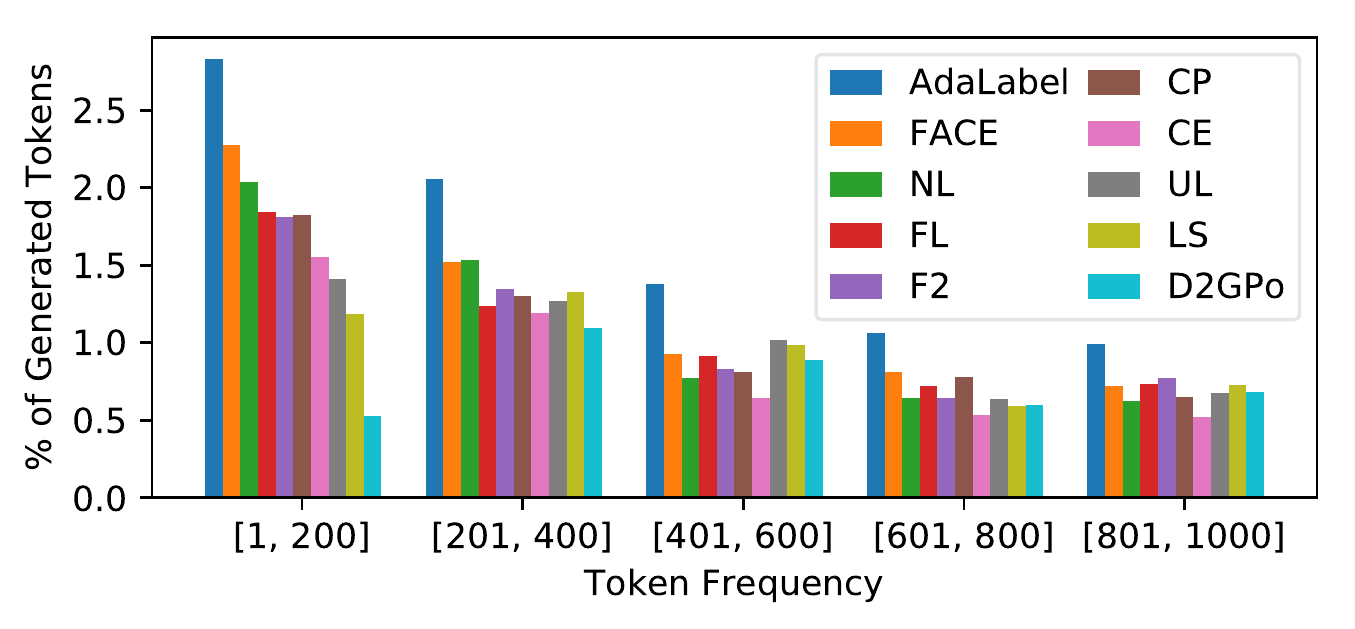}
  \caption{Ratios of low-frequency tokens in the generated responses on the OpenSubtitles dataset. Tokens in each group are determined based on the frequency on the training set.}
  \label{fig:rare_word_dist}
\end{figure}

Over-confident models produce less diversified responses because they usually under-estimate rare words. To evaluate the effectiveness of AdaLabel, we tested whether AdaLabel encourages more ``rare words'' in its generations. Specifically, the ratio of generated tokens corresponding to different token frequency bins is calculated, and the results on the OpenSubtitles dataset are shown in Figure \ref{fig:rare_word_dist}. It can be seen that AdaLabel produces more rare words in the generated responses than other baselines. Similar results are also observed on the DailyDialog dataset (see Appendix E).

\section{Conclusion}
We address the low-diversity issue of neural dialogue models by introducing an adaptive label smoothing approach, AdaLabel. In our method, the probability of each target word is estimated based on the current dialogue model's prediction, and the probabilities for these non-target words are calculated using a novel auxiliary decoder $D_a$. A target-masked attention scheme is introduced in $D_a$ to help capture forward and backward contexts. We evaluate our method on two benchmark datasets: DailyDialog and OpenSubtitles. Extensive experiments show that our method effectively alleviates the over-confidence issue and improves the diversity of the generated responses. 
As future work, we believe this method is extensible to other text generation tasks.

\section*{Acknowledgments}
This work was partly supported by the NSFC projects (Key project with No. 61936010 and regular project with No. 61876096). This work was also supported by the Guoqiang Institute of Tsinghua University, with Grant No. 2019GQG1 and 2020GQG0005. We thank Jinchao Zhang and Yao Qiu for early discussions and insightful comments of this work.

\bibliographystyle{acl_natbib}
\bibliography{acl2021}

\appendix

\begin{table*}[htp]
\centering
\small
\setlength\tabcolsep{2.2pt} 
\begin{tabular}{L{39pt}C{21pt}C{21pt}C{21pt}C{21pt}C{21pt}C{21pt}C{21pt}C{21pt}C{21pt}C{21pt}C{21pt}C{21pt}C{21pt}
C{21pt}C{21pt}C{21pt}}
\toprule
\multirow{2}{*}{Model} 
& \multicolumn{8}{c}{ DailyDialog } 
& \multicolumn{8}{c}{ OpenSubtitles } \\
\cmidrule(lr){2-9} \cmidrule(lr){10-17}
    & \multicolumn{2}{c}{Dist-1, 2} & \multicolumn{2}{c}{Ent-1, 2} & \multicolumn{1}{c}{LF} & \multicolumn{3}{c}{BLEU-2,3,4} 
    & \multicolumn{2}{c}{Dist-1, 2} & \multicolumn{2}{c}{Ent-1, 2} & \multicolumn{1}{c}{LF} & \multicolumn{3}{c}{BLEU-2,3,4} \\
    \midrule

      CE 
    &    1.79 &    8.21 &    4.19 &   5.90 &    2.57 &    4.06 &    2.49 &    1.58
    &   2.48 &   9.21 &   4.07 &   5.74 &   0.76 &   7.03 &   4.26 &   2.82 \\ 
    
      LS 
    &    1.71 &    8.01 &    4.16 &    5.89 &    2.17 &    4.13 &    2.55 &    1.65
    &   2.89 &   12.79 &   4.27 &   6.24 &   0.47 &   8.24 &   5.57 &   4.20 \\
    
      FL 
    &   2.40 &    11.37 &    4.39 &    6.35 &    4.46 &    6.01 &    3.95 &    2.75
    &   3.10 &   12.37 &   4.25 &   6.13 &   0.82 &   7.13 &   4.56 &  3.25 \\
    
      FACE 
    &    1.80 &    9.47 &    4.54 &    6.40 &    3.48 &    5.65 &    3.43 &    2.17
    &   3.12 &  12.62 &   4.47 &   6.40 &   1.02 &   5.97 &  3.63 &  2.43 \\
    
      F$^2$ 
    &    1.61 &    7.22 &    4.04 &   5.70 &    2.11 &    4.32 &    2.55 &    1.52
    &   2.89 &  10.63 &   4.03 &   5.72 &   0.89 &   6.92 &   4.27 &  2.91 \\
    
      CP 
    &   2.30 &    10.39 &    4.28 &    6.16 &    3.25 &    5.31 &    3.39 &    2.30 
    &   3.14 &   11.87 &   4.17 &   5.97 &    0.85 &   7.28 &   4.60 &   3.21 \\
    
      UL 
    &    2.42 &    11.0 &   4.40 &    6.42 &    4.55 &    7.94 &    5.26 &    3.69
    &   2.77 &   10.43 &   3.98 &   5.62 &    0.62 &    6.89 &   4.36 &   3.03 \\
    
      NL 
    &    1.61 &    7.53 &    4.19 &    6.05 &    4.02 &    7.09 &    4.41 &    2.91
    &   2.65 &  10.14 &   4.21 &   6.05 &    0.75 &    7.16 &   4.32 &  2.85\\
    
      D2GPo 
    &    1.57 &    7.83 &    4.14 &    5.91 &    2.26 &    4.47 &    2.71 &    1.71 
    &   2.06 &  10.43 &   4.15 &   6.00 &   0.12 &    7.32 &   4.69 &  3.33 \\
    
    AdaLabel 
    &   \textbf{4.25} &   \textbf{21.47} &   \textbf{4.95} &   \textbf{7.51} &   \textbf{7.68} &    \textbf{14.71} &   \textbf{11.63} &  \textbf{9.80} 
    &   \textbf{4.91} &   \textbf{21.53} &   \textbf{4.71} &   \textbf{7.08} &   \textbf{1.35} &    \textbf{8.68} &   \textbf{6.08} &  \textbf{4.68} \\
\midrule
    AdaLabel (Greedy) 
    &  3.96 &  23.53 & 5.17 &  8.00 &  8.49 &  17.42 & 13.38 &  11.01 
    &  4.78 &  22.88 & 4.96 &  7.66 &  1.47 &   9.80 &  6.48  &  4.75 \\
    
\midrule
    Human 
    &   6.59 &   37.74 &   5.67 &   8.91  &   13.7 &   N/A &   N/A &   N/A
    &   8.62 &   43.16 &   5.89 &   9.36  &   4.75 &   N/A &   N/A &   N/A  \\
    \bottomrule
\end{tabular}
\caption{
Automatic evaluation results (\%) using the beam search decoding scheme (beam size is 5).
The best results among all these beam-search-decoded models are in bold.
}
\label{tab:beam}
\label{tab:auto_res_beamsearch}
\end{table*}

\section{Implementation Details}
\label{sec:app_imp}
This appendix describes the implementation details of our model. All our experiments are implemented with python 3.7.4, PyTorch 1.7.1, and the OpenNMT package \cite{klein2017opennmt}. Training is performed on one TITAN Xp GPU. Our model's backbone is the transformer-based sequence to sequence model, the encoder and decoder each contains 6 transformer layers with 8 attention heads, and the hidden size is set to 512. The dimension of the feed-forward layer is also 512. The WordPiece tokenizer provided by BERT-base-uncased is used (the vocabulary contains 30522 tokens). The total number of parameters in our model is about 90M. The Adam optimizer is employed to train our model from random initializations with $\beta_1=0.9$, $\beta_2=0.999$, $\epsilon=1e-9$ and a learning rate of 1e-4. The batch size is set to 64 with 2 gradient accumulation so that 2 * 64 samples are used for each parameter update. The model is evaluated every 1000 steps on the validation set. We use early-stopping with patience 10, 30 for DailyDialog and OpenSubtitles, respectively. Specifically, the model stops training when the evaluation perplexity and accuracy are not increased for ``patience'' steps. The model training takes 4 hours and 3 days on DailyDialog and OpenSubtitles, respectively.

The auxiliary distribution produced by the auxiliary decoder is smoothed with the temperature scaling approach. The temperature used in this process is searched in [1, 1.5, 2]. The temperature value of 1.5 and 1.0 is used for DailyDialog, and OpenSubtitles, respectively. The hyper-parameter value of $\eta$ is set to 0.2 for all datasets. The fixed value of epsilon in our ablation model \textbf{w/o $\bm{\epsilon}$} is searched in [0.1, 0.2, 0.3, 0.4, 0.5, 0.6], and we find the value of 0.1 works best. 

\section{Baseline Implementation Details}
\label{sec:app_baseline}
This appendix contains more implementation details of our baselines. All the baselines utilize the same backbone architecture and basic hyper-parameter settings as our model (see Appendix~\ref{sec:app_imp}). The hyper-parameters specialized for each baseline is determined with the grid search based on the Dist measures on the validation set:
For \textbf{Label smoothing (LS)}, we searched the smoothing parameter in [0.05, 0.1, 0.2, 0.3, 0.4, 0.5], and found 0.1 works best on all the datasets;
For \textbf{Confidence penalty (CP)}, we searched the weight of penalty in [0.0005, 0.001, 0.01, 0.05, 0.1] and found 0.05 works best on all the datasets while ensuring the loss to be positive;
For \textbf{Focal loss (FL)}, we searched the hyper-parameter $\gamma$ in [0.1, 0.5, 1, 2, 3], and found 2 works best on all the datasets. 
For \textbf{Unlikelihood loss (UL)}, we searched the weight of penalty in [1, 10, 100, 1000], and select 1000 on all the datasets.
For \textbf{FACE}, we experiment with the Output token frequency \& PRe-weigh version, which is reported to be the best version of FACE. 
For \textbf{Negative loss (NL)}, \textbf{F$^{\bm{2}}$-softmax (F$^{\bm{2}}$)} and \textbf{Data-dependent Gaussian Prior
objective (D2GPo)}, the selection of hyper-parameters follows the author's suggestion.

\section{Automatic Evaluation Results with Other Decoding Schemes}
\label{sec:app_rand_sampling}

\begin{table*}[ht]
\centering
\small
\setlength\tabcolsep{2.2pt} 
\begin{tabular}{L{36pt}C{21pt}C{21pt}C{21pt}C{21pt}C{21pt}C{21pt}C{21pt}C{21pt}C{21pt}C{21pt}C{21pt}C{21pt}C{21pt}
C{21pt}C{21pt}C{21pt}}
\toprule
\multirow{2}{*}{Model} 
& \multicolumn{8}{c}{ DailyDialog } 
& \multicolumn{8}{c}{ OpenSubtitles } \\
\cmidrule(lr){2-9} \cmidrule(lr){10-17}
    & \multicolumn{2}{c}{Dist-1, 2} & \multicolumn{2}{c}{Ent-1, 2} & \multicolumn{1}{c}{LF} & \multicolumn{3}{c}{BLEU-2,3,4} 
    & \multicolumn{2}{c}{Dist-1, 2} & \multicolumn{2}{c}{Ent-1, 2} & \multicolumn{1}{c}{LF} & \multicolumn{3}{c}{BLEU-2,3,4} \\
    \midrule

      CE 
    &    2.22 &    19.05 &    5.07 &    7.87 &    4.09 &    6.78 &    3.29 &    1.61
    &   3.78 &  20.58 &   5.07 &   7.97 &   1.23 &   5.94 &   2.84 &   1.46 \\ 
    
      LS 
    &    1.95 &    17.74 &    5.02 &    7.82 &    3.69 &    7.08 &   3.50 &    1.77
    &   3.46 &   21.27 &   5.10 &   8.12 &   0.78 &   6.15 &   \textbf{3.16} &   \textbf{1.85} \\
    
      FL 
    &    2.71 &    20.98 &    5.19 &    8.17 &    6.44 &    8.09 &    4.13 &    2.24
    &   3.82 &   22.14 &   5.15 &   8.25 &   1.27 &   5.34 &   2.54 &  1.34 \\
    
      FACE 
    &    2.29 &    21.14 &    5.36 &    8.3 &    5.73 &    7.07 &    3.47 &    1.82
    &   4.25 &  23.95 &   5.30 &   8.37 &   1.51 &   5.34 &  2.54 &  1.33 \\
    
      F$^2$ 
    &    2.16 &    19.33 &    5.04 &    7.85 &    3.97 &    6.31 &    3.12 &    1.58
    &   4.10 &  22.53 &   5.13 &   8.11 &   1.32 &   5.27 &   2.51 &  1.31 \\
    
      CP 
    &    3.16 &    22.38 &    5.11 &    7.96 &    6.01 &    8.11 &    4.38 &    2.50 
    &   4.06 &   22.62 &   5.13 &   8.14 &    1.33 &   6.00 &   2.94 &   1.52 \\
    
      UL 
    &    2.92 &    20.81 &    5.12 &    7.99 &    6.44 & \textbf{9.36} &    \textbf{5.13} &    \textbf{3.00}
    &   3.74 &   20.97 &   5.01 &   7.94 &    1.00 &    6.01 &   2.99 &   1.64 \\
    
      NL 
    &    2.39 &    18.35 &    4.99 &    7.79 &    5.72 &    8.71 &    4.64 &    2.63
    &   3.57 &   20.36 &   5.05 &   7.97 &    1.06 &    5.84 &   2.86 &   1.46 \\
    
      D2GPo 
    &    1.75 &    17.09 &   5.00 &    7.81 &   3.40 &    7.45 &    3.73 &    1.97 
    &   2.74 &  19.21 &   5.00 &   7.97 &   0.36 &    \textbf{6.32} &   3.15 &  1.72 \\
    
    AdaLabel 
    &   \textbf{4.11} &   \textbf{32.65} &   \textbf{5.58} &   \textbf{8.93} &   \textbf{10.99}  &    8.87 &    4.84 &    2.90
    &   \textbf{4.78} &   \textbf{29.58} &   \textbf{5.43} &   \textbf{8.78} &   \textbf{1.53} &    5.12 &   2.32 &  1.19 \\
    
\midrule
    AdaLabel (Greedy) 
    &  3.96 &  23.53 & 5.17 &  8.00 &  8.49 &  17.42 & 13.38 &  11.01 
    &  4.78 &  22.88 & 4.96 &  7.66 &  1.47 &   9.80 &  6.48  &  4.75 \\
    
\midrule
    Human 
    &   6.59 &   37.74 &   5.67 &   8.91  &   13.7 &   N/A &   N/A &   N/A
    &   8.62 &   43.16 &   5.89 &   9.36  &   4.75 &   N/A &   N/A &   N/A  \\
    \bottomrule
\end{tabular}
\caption{
Automatic evaluation results (\%) using the top-k sampling decoding scheme ($k=10$).
The best results among all these top-k-decoded models are in bold.
}
\label{tab:topk}
\end{table*}

This appendix reports our model's automatic evaluation results and all the baselines when different decoding schemes are used. Specifically, Table \ref{tab:beam} shows the results for the beam search decoding scheme (beam size of 5), and Table \ref{tab:topk} shows the results when the top-K decoding scheme ($k=10$) is used. Note that for the F$^2$-softmax, we use the decoupled top-k sampling as the authors suggested.

As can be seen from Table \ref{tab:beam} and \ref{tab:topk}, our method outperforms all the baselines on the diversity-related scores (i.e., Dist, Ent, and LF) by a large margin. This indicates that our method can produce more diverse responses even with the stochastic based decoding scheme. 

We also include the results of AdaLabel when the greedy decoding scheme is used in Table \ref{tab:beam} and Table \ref{tab:topk} (the second line from the bottom). It is interesting to see that the greedily decoded responses from AdaLabel are more diverse than some baselines that are decoded using the sampling scheme (see Table \ref{tab:topk}). Moreover, our model AdaLabel with the greedy decoding scheme achieves the best BLEU among all the baselines on both datasets.

\section{Prediction Confidence}
\label{sec:app_pred_conf}

This appendix reports the prediction confidence scores assigned by each model to high-frequency words. Specifically, words occupying the top 40\% of the frequency mass in the training set of each dataset are regarded as high-frequency words.

\begin{figure*}[!ht]
  \centering
  \includegraphics[width=430px]{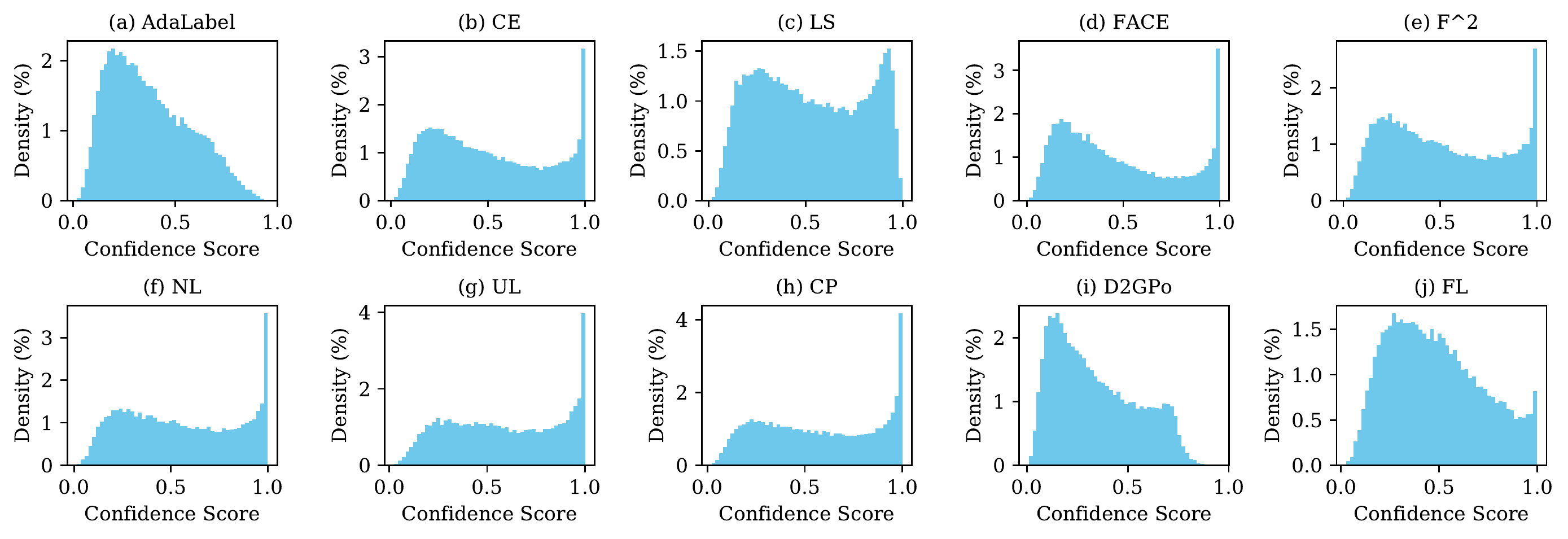}
  \caption{Confidence score distributions for high-frequency words on the DailyDialog dataset. Words occupying the top 40\% of the frequency mass in the training set of DailyDialog are regarded as high-frequency words.}
  \label{fig:conf_dist_daily}
\end{figure*}

\begin{figure*}[!ht]
  \centering
  \includegraphics[width=430px]{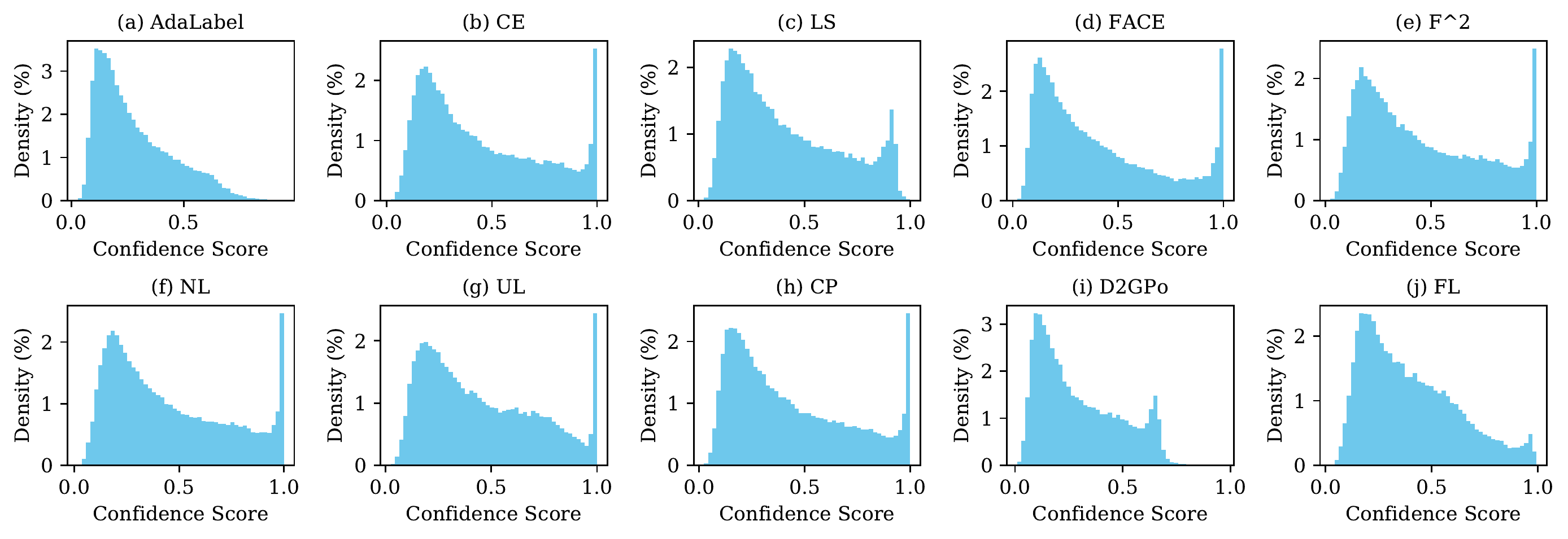}
  \caption{Confidence score distributions for high-frequency words on the OpenSubtitles dataset. Words occupying the top 40\% of the frequency mass in the training set of OpenSubtitles are regarded as high-frequency words.}
  \label{fig:conf_dist_ost}
\end{figure*}

Figure \ref{fig:conf_dist_daily} shows the results of our model and all the baselines on the DailyDialog dataset. Figure \ref{fig:conf_dist_ost} shows the results of our model and all the baselines on the OpenSubtitles dataset. It can be seen that most of our baselines assign extremely high confidence scores (nearly 1.0) to these high-frequency words, and thus resulting in a spike of high confidence scores in the plotted distribution. Our model outperforms all the baselines in avoiding assigning extremely high confidence scores to these high-frequency words.

\section{Predicted Rare Word Distribution on DailyDialog}
\label{sec:app_rare_dist}

\begin{figure}[!t]
  \centering
  \includegraphics[width=210px]{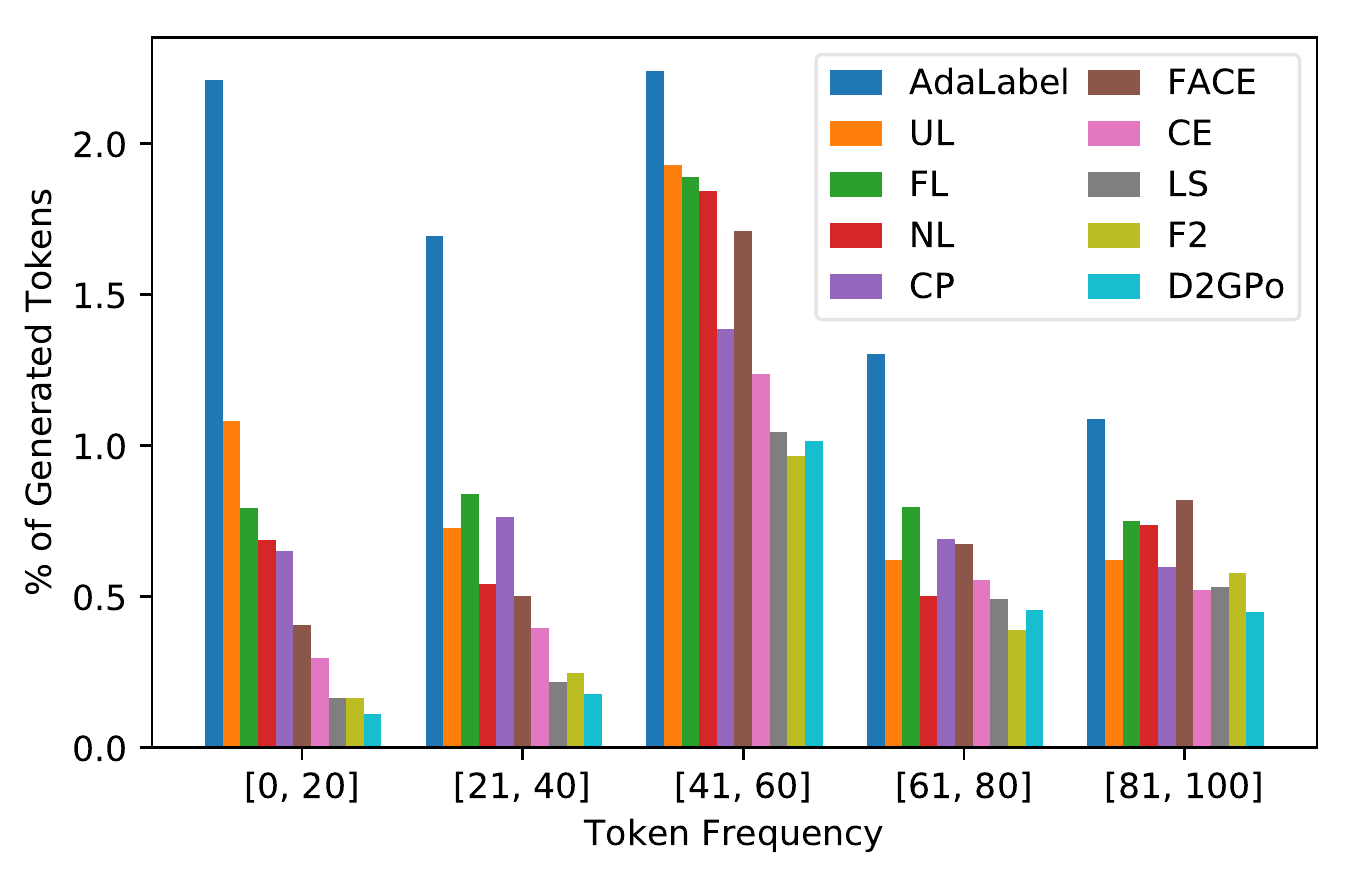}
  \caption{Ratios of low-frequency tokens in the generated responses on the DailyDialog dataset. Tokens in each group are determined based on the frequency on the training set.}
  \label{fig:rare_word_dist_daily}
\end{figure}

This appendix shows the distribution of rare words in the generated responses on the DailyDialog dataset (see Figure \ref{fig:rare_word_dist_daily}). It can be seen that more ``rare words'' are predicted by our method on the DailyDialog dataset. This observation is in line with the results on the OpenSubtitles dataset as reported in Section 5.3.

\section{Use BERT Model to Obtain $\bm{v}$}
\label{sec:bert_model}

\begin{table}[t]
\small
\setlength\tabcolsep{1pt}
\centering
  \begin{tabular}{lC{21pt}C{21pt}C{21pt}C{21pt}C{21pt}C{21pt}C{21pt}}
  \toprule
    Model &  \multicolumn{2}{c}{BLEU-3,4}& \multicolumn{2}{c}{Dist-1,2} & \multicolumn{2}{c}{Ent-1,2} & LF   \\
  \midrule
    1. CMLM          &   6.18 &   4.09 &   2.20 &   11.83 &   4.59 &   6.79 &   4.62 \\
    2. CMLM+$\varepsilon$ &   9.36 &   7.31 &   3.78 &   21.05 &   4.96 &   7.61 &   6.88 \\
    3. CMLM+$\varepsilon$+$D_a$ &   11.6 &   9.34 &   3.67 &   20.97 &   5.02 &   7.71 &   7.28 \\

  \midrule
    AdaLabel &  \textbf{13.38} &   \textbf{11.01}     &    \textbf{3.96} &   \textbf{23.53} &   \textbf{5.17} &   \textbf{8.00} &   \textbf{8.49} \\
  \bottomrule
  \end{tabular}
\caption{
Ablation study results based on BERT on DailyDialog (\%). 
}
\label{tab:extra ablation}
\end{table}

This appendix provides more experiment results comparing to the CMLM model \cite{chen2020distilling}: \textbf{1). CMLM} exactly follows the setting of \citet{chen2020distilling}, i.e., the teacher distribution produced by the BERT model is merged with the one-hot distribution using a fixed $\varepsilon$. \textbf{2). CMLM+$\varepsilon$} adaptively adjust the value of $\varepsilon$ using Eq. 6 in our paper. \textbf{3). CMLM+$\varepsilon$+$D_a$} add an additional training task to optimize the auxiliary decoder $D_a$ on the basis of \textbf{CMLM+$\varepsilon$}. It is expected that optimizing $D_a$ help our dialogue encoder to capture better representations. The trained $D_a$ is not used in the training and inference phase of our dialogue model. Note that the last model \textbf{CMLM+$\varepsilon$+$D_a$} is the same with our ablation model \textbf{6. BERT} as reported in our paper.

As can be seen Table~\ref{tab:extra ablation}, our approach to adaptively change $\varepsilon$ helps to produce better dialogue responses, and the training of $D_a$ helps our dialogue encoder to learn better representations.

\section{Case study}
\label{sec:app_case}
We sampled some generated cases on the DailyDialog and OpenSubtitles dataset. The results of our model and some competitive baselines are shown in Table \ref{tab: cases DailyDialog} and Table \ref{tab: cases OpenSubtitles}. It can be seen that the responses generated by our method are coherent to the context and contain richer contents. Moreover, our model also produces more rare words that make our response more diverse.

\begin{table}[ht]
\centering
\small
\setlength\tabcolsep{2.0pt} 

    \begin{tabular}{ll}
    \toprule
     & Dialogue  \\
    \midrule
    
    Context1 &  Tell me a little bit about yourself , please . \\
    \midrule
    
    UL       &  I am sorry , sir . I have been in a foreign \\
             &  currency company for the company . \\
     
    D2GPo    &  I ' m afraid i ca n ' t . I ' m a little \\
             &  nervous about it . \\
     
    FACE     &  Yes , i am in the mood for a few days ago . \\

    AdaLabel   &  I am studying in the international trade \\
             &  company . I am studying english in my  company . \\
     
    
     
     

    
    \midrule
    
    Context1  &  Oh , they look nice . may i try that black one on ? \\
    Context2  &  Of course . the fitting room is right over there . \\
    \midrule
              
    UL        &  I think you ' re right . i think you ' ll have a try . \\
     
    D2GPo     &  I think it ' s a good idea . \\
     
    FACE      &  How much is it ? \\

    AdaLabel    &  I bought this one . it ' s the latest fashion . \\

    \midrule
    
    Context1  &  Well , we don ' t have any pre - existing  \\
              &  illnesses or conditions , so we should be fine .\\
    \midrule

    UL        &  And most people are n ' t so quiet and \\
              &  interviewing .\\
     
    D2GPo     &  And the company company are a good job in \\
              &  the company . \\
     
    FACE      &  And your company insurance , or workshops  \\
              &  are booked in a private company ? \\

    AdaLabel    &  I wish our company or country provided us  \\
              &  with healthcare . \\
    
    \bottomrule
    \end{tabular}
\caption{
Cases on DailyDialog.
}
\smallskip
\label{tab: cases DailyDialog}
\end{table}

\begin{table}[ht]
\centering
\small
\setlength\tabcolsep{2.0pt} 

    \begin{tabular}{ll}
    \toprule
     & Dialogue  \\
    \midrule
    
    Context  &  You ' re going to love this baby . \\
    \midrule
    LS       &  I ' m going to get you . \\
     
    D2GPo    &  I ' m going to go to the bathroom . \\
     
    FACE     &  You ' re going to love this baby . \\

    AdaLabel   &  You ' re going to be a good father . \\
    
    \midrule
    
    Context  &  Mr. Matlack can ' t offend . \\
    \midrule
             
    LS       &  This is our official business . \\
     
    D2GPo    &  This is the most important thing . \\
     
    FACE     &  It ' s a pleasure to meet you , mr .\\
             &  President . \\

    AdaLabel   &  Timothy Matlack is the official scribe\\
             &  of the continental congress . \\
    
    \midrule
    
    Context  &  Brave men discover they are really cowards . \\
    \midrule
    LS       &  I ' m a samurai . \\
     
    D2GPo    &  I ' ve never seen such a beast in my life . \\
     
    FACE     &  They are not afraid of you . \\

    AdaLabel   &  Father , you ' re a coward ! \\
    
    \midrule
    
    Context  &  I found this whole operation strange . \\
    \midrule
    LS       &  I ' m not going to let you go . \\
     
    D2GPo    &  I ' m not sure i ' m going to do it . \\
     
    FACE     &  I ' m sorry , but i ' m not sure i can handle it .\\

    AdaLabel   &  It ' s an incredibly delicate system . \\
    
    \midrule
    
    Context  &  If they make it , they ' re clear into a safe zone\\
             &  where they can get medical supplies and food .\\
    \midrule
    LS       &  We ' il get them to the safe . \\
     
    D2GPo    &  We ' il have to get back to the hotel . \\
     
    FACE     &  They ' re gon na get us out of here . \\

    AdaLabel   &  So we can use it as a safe field . \\
   
    \bottomrule
    \end{tabular}
\caption{
Cases on OpenSubtitles.
}
\smallskip
\label{tab: cases OpenSubtitles}
\end{table}

\end{document}